\documentclass[letterpaper, 10 pt, conference]{ieeeconf}  

\IEEEoverridecommandlockouts 
\usepackage[english]{babel}
\usepackage{url}
\usepackage{subfigure}

\usepackage{graphicx} 
\usepackage{amsmath}
\usepackage{amssymb}
\usepackage{textcomp} 
\usepackage{acronym}
\usepackage{algorithm}

\DeclareMathOperator{\betP}{betP}

\usepackage[colorinlistoftodos]{todonotes}

\newcommand{\MyTitle}{Enhancing Mobile Object Classification \texorpdfstring{\linebreak[4]}{} Using Geo-referenced Maps and Evidential Grids}
\newcommand{\MyKeywords}{dynamic fusion, geo-referenced maps, mobile perception, prior knowledge, evidential occupancy grid, autonomous vehicle}

\usepackage{hyperref}
\hypersetup{
    bookmarksnumbered=true,
    bookmarksopen=true,
    bookmarksopenlevel=2,
    colorlinks=true, 
    linkcolor=black,
    citecolor=black, 
    filecolor=black,
    urlcolor=black,
        pdfinfo={
            Title={\MyTitle},
            Keywords={\MyKeywords},
            Author={Marek Kurdej, Julien Moras, V\'{e}ronique Cherfaoui, Philippe Bonnifait},
        }
}

\newcommand{\email}[1]{%
    \href{mailto:#1}{\texorpdfstring{\texttt{{#1}}}{#1}}%
}
\newcommand{\quotemarks}[1]{``#1''}

\newcommand{\oppause}{\:}
\newcommand{\opconj}{\oppause\text{\textcircled{\scriptsize{$\cap$}}}\oppause}
\newcommand{\opdempster}{\oppause\oplus\oppause}
\newcommand{\opdisj}{\oppause\text{\textcircled{\scriptsize{$\cup$}}}\oppause}

\def\padfivezeroes#1{%
    \ifnum#1<10000 0\fi
    \ifnum#1<1000 0\fi
    \ifnum#1<100 0\fi
    \ifnum#1<10 0\fi
    \number#1%
}

\acrodef{SLAM}{Simultaneous Localisation and Mapping}
\acrodef{IGN}{French National Geographic Institute}
\acrodef{GG}{\texttt{GISGrid}}
\acrodef{PG}{\texttt{PerceptionGrid}}
\acrodef{SG}{\texttt{SensorGrid}}
\acrodef{FOD}{frame of discernment}
\acrodef{DST}{Dempster--Shafer theory}
\acrodef{bba}{basic belief assignment}
\title{\LARGE{\textbf{\MyTitle}}}
\author{Marek Kurdej, \and Julien Moras, \and V\'{e}ronique Cherfaoui, \and Philippe Bonnifait
\thanks{
    *
    Authors are with
    UMR CNRS 7253 \href{http://www.hds.utc.fr}{Heudiasyc}
    \href{http://www.utc.fr}{University of Technology of Compi\`{e}gne}, France.
    E-mail: \email{firstname.surname@hds.utc.fr}
}
}
\begin{document}
\maketitle

\begin{abstract}

Evidential grids have recently shown interesting properties for mobile object perception.
Evidential grids are a generalisation of Bayesian occupancy grids using Dempster--Shafer theory.
In particular, these grids can handle efficiently partial information.
The novelty of this article is to propose a perception scheme enhanced by geo-referenced maps
    used as an additional source of information,
    which is fused with a sensor grid.
The paper presents the key stages of such a data fusion process.
An adaptation of conjunctive combination rule is presented to refine the analysis of the conflicting information.
The method uses temporal accumulation to make the distinction between stationary and mobile objects,
    and applies contextual discounting for modelling information obsolescence.
As a result, the method is able to better characterise the occupied cells by differentiating, for instance, moving objects, parked cars, urban infrastructure and buildings.
Experiments carried out on real-world data illustrate the benefits of such an approach.

\keywords
\MyKeywords

\end{abstract}

\section{Introduction}
\label{sec:introduction}


Autonomous driving has been an important challenge in recent years.
Navigation and precise localisation aside, environment perception is an important on-board system of a self-driven vehicle.
The level of difficulty in autonomous driving increases in urban environments, where a good scene understanding makes the perception subsystem crucial.
There are several reasons that make cities a demanding environment.
Poor satellite visibility deteriorates the precision of GPS positioning.
Vehicle trajectories are hard to predict due to high variation in speed and direction.
Also, the sheer number of mobile objects poses a problem, e.g. for tracking algorithms.

On the other hand, more and more detailed and precise geographic databases become available.
This source of information has not been well examined yet, hence our approach of incorporating prior knowledge from digital maps in order to improve perception scheme.
A substantial amount of research has focused on the mapping problem for autonomous vehicles,
e.g. \ac{SLAM} approach \cite{Thrun2005}, but the use of maps for perception is still understudied.


In this article, we propose a new perception scheme for intelligent vehicles.
The information fusion method is based on Dempster--Shafer theory of evidence \cite{Shafer1976}.
The principal innovation of the method is the use of meta-knowledge obtained from a digital map.
The map is considered as an additional source of information on a par with other sources, e.g. sensors.
We show the advantage of including prior knowledge into an embedded perception system of an autonomous car.
To model the vehicle environment, our approach uses multiple 2D evidential occupancy grids described in \cite{Pagac1998a}.
Originally, occupancy grids containing probabilistic information were proposed in \cite{Elfes1989}.

Our method aims to model complex vehicle environment, so that it can be used as a robust world representation for other systems, such as navigation.
We want to detect mobile and static objects and distinguish stopped and moving objects.
The objective of the proposed scheme is to model the free and navigable space as well.


This paper describes a robust and unified approach to a variety of problems in spatial representation using the Dempster--Shafer theory of evidence.
The theory of evidence was not combined with occupancy grids until recently to build environment maps for robot perception \cite{Pagac1998a}.
Only recent works take advantage of the theory of evidence in the context of mobile perception \cite{Moras2011}.
There is also some research on efficient probabilistic and 3-dimensional occupancy grids \cite{Wurm2010a}.
Some authors have also used a laser range scanner as an exteroceptive source of information \cite{Moras2011}. 
Some works use 3D city model as a source of prior knowledge for localisation and vision-based perception \cite{Dawood2011},
    whereas our method uses maps for scene understanding.
Geodata are also successfully used for mobile navigation \cite{Hentschel2010}.


This article is organised as follows.
Section~\ref{sec:theory} gives necessary theoretical background of the Dempster--Shafer theory of evidence.
In section~\ref{sec:method}, we describe the details of the proposed method, starting with the description of needed data and the purpose of each grid.
Further, details on the information fusion are given.
Data-dependent computation which are not in the heart of the method are described in section~\ref{sec:setup}.
Section~\ref{sec:results} presents the results obtained with real-world data.
Finally, section \ref{sec:conclusion} concludes the paper and presents ideas for future work.

\section{\acl{DST} of evidence}
\label{sec:theory}

The \acf{DST} is a mathematical theory specially adapted to model the uncertainty and the lack of information introduced by Dempster and further developed by Shafer \cite{Shafer1976}.
\ac{DST} generalises the theory of probability, the theory of possibilities and the theory of fuzzy sets.
In the \acf{DST}, a set $\Omega = \omega_1, \ldots, \omega_n $ of mutually exclusive propositions is called the \acf{FOD}.
In case of closed-world hypothesis, the \ac{FOD} presents also an exhaustive set.
Main difference in comparison to the theory of probability is the fact
    that the mass of evidence is attributed not only to single hypotheses (singletons),
    but to any subset of the \ac{FOD}, including an empty set.

As stated in the previous paragraph, beliefs about some piece of evidence are modelled by the attribution of mass to the corresponding set.
This attribution of mass, called a \acf{bba}, or a mass function, is defined as a mapping:
\begin{align}
    m(\cdot) \textbf{ : } 2^{\Omega} & \mapsto [0, 1]
    \\
    \sum\limits_{A \subseteq \Omega} m(A) &= 1
    \\
    m(\emptyset) &= 0
\end{align}

In order to combine various information sources in the \ac{DST}, there are many rules of combination.
Combined mass functions have to be defined on the same \ac{FOD} $\Omega$ or transform to a common frame using refining functions.
A \emph{refining} is defined as a one-to-many mapping from $\Omega_1$ to $\Omega_2$. 
\begin{align}
    r \textbf{ : } 2^{\Omega_{1}} &\mapsto 2^{\Omega_{2}} \setminus \emptyset
    \\
    r(\omega) &\neq \emptyset
        & \forall \omega \in \Omega_1
    \\
    \bigcup\limits_{\omega \in \Omega_1} r(\omega) &=
        \Omega_2
    \\
    r(A) &= \bigcup_{\omega \in A} r(\omega)
\end{align}
The \acl{FOD} $\Omega_2$ is then called the \emph{refinement} of $\Omega_1$,
    and $\Omega_1$ is the \emph{coarsening} of the $\Omega_2$.

When combined pieces of evidence expressed by \ac{bba}s are independent and both are reliable,
    then the conjunctive rule and Dempster's combination rule are commonly used.
In the case when the sources are independent, but only one of them is judged reliable, a disjunctive rule is used.

In the following, let us suppose that $m_1, m_2$ are \ac{bba}s.
Then, the conjunctive rule of combination denoted by $\opconj$ is defined as follows:
\begin{align}
    (m_1 \opconj m_2)(A) &=
        \sum\limits_{A = B \cap C} m_1(B) \cdot m_2(C)
\end{align}

The combination using the conjunctive rule can generate the mass on the empty set $m(\emptyset)$.
This mass can be interpreted as the conflict measure between the combined sources.
Therefore, a normalised version of conjunctive rule, called Dempster's conjunctive rule and noted $\opdempster$ was defined:
\begin{align}
    (m_1 \opdempster m_2)(A) &= \frac{(m_1 \opconj m_2)(A)}{1 - K}
    \\
    (m_1 \opdempster m_2)(\emptyset) &= 0
    \\
    K &= (m_1 \opconj m_2)(\emptyset)
\end{align}

The disjunctive rule of combination, noted $\opdisj$ is defined as follows:
\begin{align}
    (m_1 \opdisj m_2)(A) &=
        \sum\limits_{A = B \cup C} m_1(B) \cdot m_2(C)
\end{align}


\begin{table}
    \centering
    \begin{tabular}{||c||r|r|r|r||}
        \hline \hline
            & $\emptyset$ & $a$ & $b$ & $\Omega = \{ a, b \}$ \\
        \hline \hline
        $m_1$ & 0 & 0.2 & 0.6 & 0.2 \\
        \hline
        $m_2$ & 0 & 0.7 & 0.1 & 0.2 \\
        \hline
        $m_1 \opconj m_2 $ & 0.44 & 0.34 & 0.18 & 0.04 \\
        \hline
        $m_1 \opdempster m_2 $ & 0 & 0.61 & 0.32 & 0.07 \\
        \hline
        $m_1 \opdisj m_2 $ & 0 & 0.14 & 0.06 & 0.8 \\
        \hline
        ${}^{\alpha} m_1 $ & 0 & 0.18 & 0.54 & 0.28 \\
        \hline
        $ \betP_1 $ & 0 & 0.3 & 0.7 & 1 \\
        \hline \hline
    \end{tabular}
    \caption{Example of fusion rules, discounting with $ \alpha = 0.1 $ and pignistic probability.}
    \label{tab:fusion-rules}
\end{table}

In the \ac{DST}, a discounting operation is used in order to, e.g. model information ageing.
Discounting in its basic form requires to set a discounting factor $\alpha$ and is defined as:
\begin{align}
    {}^\alpha m(A) &=
        (1 - \alpha) \cdot m(A)
        & \forall A \subsetneq \Omega
    \\
    {}^\alpha m(\Omega) &= 
        (1 - \alpha) \cdot m(\Omega) + \alpha
\end{align}

Decision making in \ac{DST} creates sometimes a necessity of transforming a mass function into a probability function \cite{Smets2005}.
Smets and Kennes proposed so called \emph{pignistic transformation} in \cite{Smets1994}.
Pignistic probability $\betP$ has been defined as:
\begin{align}
    \betP(B) = \sum\limits_{A \in \Omega} m(A) \cdot \frac{|B \cap A|}{|A|}
\end{align}
where $|A|$ is the cardinality of the set $A$.

Table \ref{tab:fusion-rules} presents an example of different combination rules, pignistic transform and discounting operation.

\section{Multi-grid fusion approach}
\label{sec:method}

\begin{figure*}[htb!]
    \centering
    \includegraphics[width=\textwidth]{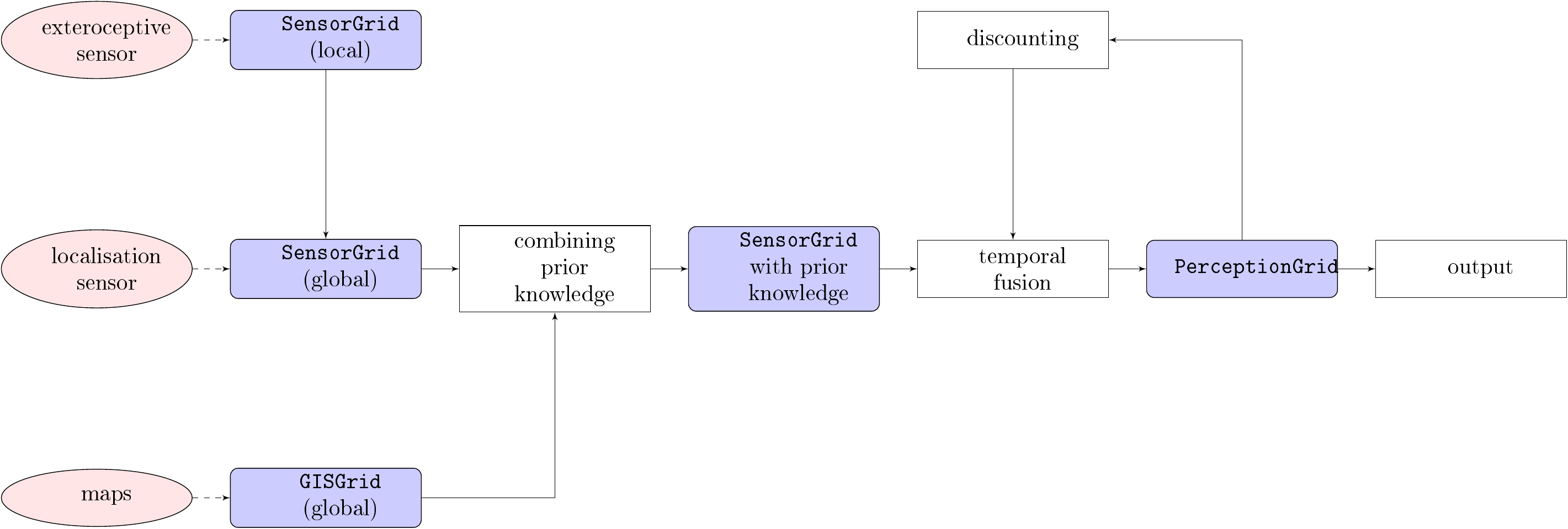}
    \caption{Method overview.}
    \label{fig:fusion-general}
\end{figure*}

This section presents the proposed perception schemes.
We use three evidential occupancy grids to model prior information, sensor acquisition and perception result.
The grid construction method is described in section~\ref{grids}.
We detail all data processing steps in section~\ref{fusion}.
Figure~\ref{fig:fusion-general} presents a general overview of our approach.
Following sections correspond to different blocks of this diagram.

\subsection{Heterogeneous data sources}
\label{sources}

There are three sources in our perception system: vehicle pose, exteroceptive acquisition data and vector maps.
Figure~\ref{fig:fusion-general} illustrates all system inputs.
The proposed approach is based on the hypothesis that all these information sources are available.
Other hypotheses on the input data are done.
Firstly, a globally referenced vehicle pose is needed to situate the system in the environment.
The pose provided by a proprioceptive sensor should be reliable, integrate and as precise as possible.
It is assumed that the pose reflects closely the real state of the vehicle.
Secondly, an exteroceptive sensor supplies a partial view of the environment.
This sensor should be able to at least distinguish free and occupied space, and model it in 2D $x, y$ or 3D $x, y, z$ coordinates.
The coordinates can be globally referenced or relative to the vehicle.
A typical exteroceptive sensor capable of satisfying this assumption is a Lidar (laser range scanner), radar, or a stereo camera system.
Lastly, our method tries to exploit at large the information contained in vector maps,
    so we assume that the maps are sufficiently rich and contain valuable accurate data.
Typically, map data should contain information on the location of buildings and the model of road surface.

\subsection{Occupancy grids}
\label{grids}

An occupancy grid models the world using a tessellated representation of spatial information.
In general, it is a multidimensional spatial lattice with cells storing some stochastic information.
In our case, each cell representing a box (a part of environment) $X\times Y$ where $X=\left[x_{-},\, x_{+}\right]$, $Y=\left[y_{-},\, y_{+}\right]$ stores a mass function.

\subsubsection{\acf{SG}}
\label{sensor-grid}

In order to process the exteroceptive sensor data, an evidential occupancy grid is computed when a new acquisition arrives, this grid is called \acl{SG}. Each cell of this grid stores a mass function on the \ac{FOD} $\Omega_{SG} = \left\{ F,O\right\} $, where $F$ refers to the free space and $O$ -- to the occupied space. The basic belief assignment reflects the sensor model.

\subsubsection{\acf{PG}}
\label{perception-grid}

To store the results of information fusion, an occupancy grid PG has been introduced with a \ac{FOD} $\Omega_{PG}=\left\{ F,\, I,\, U,\, S,\, M\right\} $.
The choice of such a \ac{FOD} is directly coupled with the objectives that we try to achieve.
Respective classes represent:
    free space $F$,
    mapped infrastructure (buildings) $I$,
    unmapped infrastructure $U$,
    temporarily stopped objects $S$
    and mobile moving $M$ objects.
$\Omega_{PG}$ is a common frame used for information fusion.
By using \ac{PG} as a cumulative information storage, we are not obliged to store preceding \acl{SG}s.

\subsubsection{\acf{GG}} 
\label{gis-grid}

This grid allows us to perform a contextual information fusion incorporating some meta-knowledge about the environment.
\acl{GG} uses the same frame of discernment $\Omega_{PG}$ as \acl{PG}.
The grid can be obtained, for instance, by projection of map data, buildings and roads, onto a 2D grid with global coordinates.
However, the exact method of creating the \ac{GG} depends on available GIS information.
Section~\ref{gis-grid-construction} presents how the \ac{GG} was constructed.

\subsection{Combining prior knowledge}
\label{prior}

In our method, prior information contained in maps serves to ameliorate the perception scheme.
We have chosen to combine the prior knowledge with the sensor data of the \acl{SG}.
However, the \acl{DST} does not allow to combine sources with different frames of discernment.
The frame of discernment $\Omega_{SG}$ is distinct from $\Omega_{PG}$ used in \acl{GG}.
Hence, we are obliged to find a common frame for both sources.
In order to enable the fusion of \acf{SG} and \acf{GG}, we define a refining:
\begin{align}
    r_{SG}\textbf{ : } 2^{\Omega_{SG}} &\mapsto 2^{\Omega_{PG}}
    \\
    r_{SG}\left(\left\{ F\right\} \right) &= \left\{ F\right\}
    \\
    r_{SG}\left(\left\{ O\right\} \right) &= \left\{ I,U,S,M\right\}
    \\
    r_{SG}(A) &= \bigcup_{\theta\in A}r_{SG}(\theta)
\end{align}

Refining $r$ allows us to combine prior knowledge included in \acl{GG} with instantaneous grid obtained from sensor(s).

The refined mass function can be expressed as:
\begin{align}
    m_{SG} ^ {\Omega_{PG}}\left(r_{SG}\left(A\right)\right) &=
        m_{SG}^{\Omega_{SG}}\left(A\right)
        & \forall A\subseteq\Omega_{SG}
\end{align}

Then, Dempster's rule described in section~\ref{sec:theory} is applied in order to exploit the prior information
included in \ac{GG}:
\begin{align}
    m'{}_{SG,\, t}^{\Omega_{PG}} &=
        m_{SG,\, t}^{\Omega_{PG}}\opdempster m_{GG}^{\Omega_{PG}}
\end{align}

We have chosen to use the Dempster's rule of combination, since the GIS data and the sensor data are independent.
Besides, we suppose that both sources are reliable, even if errors are possible.
In the end of this stage, we obtain a grid being combination of the sensor data, \acl{SG}, with the prior knowledge from \acl{GG}.

\subsection{Temporal fusion}
\label{fusion}

The role of the fusion operation is to combine current sensor acquisition with preceding perception result.
The sensor acquisition input is already combined with prior information as described in preceding paragraphs.
We propose to exploit dynamic characteristics of the scene
    by analysing produced conflict masses.
As the preceding perception result \acl{PG} is partially out-of-date at the moment of fusion,
    the contextual discounting operation is employed to model this phenomena.
Moreover, a counter of occupancy has been introduced and a mass function specialisation is performed
    to distinguish mobile, but temporarily stopped objects.

\subsubsection{Computing conflict masses}
\label{conflict}

To distinguish between two types of conflict
    which arise from the fact that the environment is dynamic,
    the idea from \cite{Moras2010} is used.
$\emptyset_{FO}$ denotes the conflict induced when a free cell in PG is fused with an occupied cell in SG.
Similarly, $\emptyset_{OF}$ indicates the conflicted mass caused by an occupied cell
in \ac{PG} fused with a free cell in \ac{SG}.

Conflict masses are calculated using the formulas:
\begin{align}
    m_{PG,\, t} \left(\emptyset_{OF}\right) &=
        m_{PG,\, t-1}\left(O\right)\cdot m_{SG,\, t}\left(F\right)
    \\
    m_{PG,\, t}\left(\emptyset_{FO}\right) &=
        m_{PG,\, t-1}\left(F\right)\cdot m_{SG,\, t}\left(O\right)
\end{align}
where $m(O)=\sum\limits _{A}m(A),\;\forall A\subseteq\left\{ I,U,S,M\right\} $.
In an error-free case, these conflicts represent, respectively, the disappearance and the appearance of an object.

\subsubsection{\acl{PG} specialisation using an accumulator}
\label{specialisation}

Mobile object detection is an important issue in dynamic environments.
We propose the introduction of an accumulator $\zeta$ in each cell in order to include temporal information on the cell occupancy.
For this purpose, incrementation and decrementation steps $\delta_{inc}\in[0,1]$,
$\delta_{dec}\in[0,1]$, as well as threshold values $\gamma_{O}$,
$\gamma_{\emptyset}$ have been defined.

\begin{align}
    \zeta^{(t)} & =\min\left(1,\,\zeta^{(t-1)}+\delta_{inc}\right)
        \\
        & \text{if }m_{PG}(O)\geq\gamma_{O}
        \nonumber
        \\
        & \text{ and } m_{PG}\left(\emptyset_{FO}\right)+m_{PG}\left(\emptyset_{OF}\right)\leq\gamma_{\emptyset}
        \nonumber
    \\
    \zeta^{(t)} & = \max\left(0,\,\zeta^{(t-1)}-\delta_{dec}\right)
        \\
        & \text{if } m_{PG} \left( \emptyset_{FO} \right) + m_{PG} \left(\emptyset_{OF}\right)>\gamma_{\emptyset}
        \nonumber
    \\
    \zeta^{(t)} & = \zeta^{(t-1)}
        \\
        & \text{ otherwise}
\end{align}

Using $\zeta$ values, we impose a specialisation of mass functions
in \ac{PG} using the equation:
\begin{align}
    m'{}_{PG,\, t} & \left(A\right)=S(A,B)\cdot m_{PG,\, t}(B)
\end{align}
where specialisation matrix $S(\cdot,\cdot)$ is defined as:

\begin{align}
    S(A\backslash\left\{ M\right\} ,\, A) & =\zeta & \forall A\subseteq\Omega_{PG} \text{ and } \left\{ M\right\} \in A\nonumber \\
    S(A,\, A) & =1-\zeta & \forall A\subseteq\Omega_{PG} \text{ and } \left\{ M\right\} \in A\nonumber \\
    S(A,\, A) & =1 & \forall A\subseteq\Omega_{PG} \text{ and } \left\{ M\right\} \notin A\nonumber \\
    S(\cdot,\,\cdot) & =0 & \text{otherwise}
\end{align}

The idea behind the specialisation matrix and the accumulator is that the mass attributed to set $N,S,M$ or $S,M$ will be transferred to set $N,S$ or $S$, respectively.
The transferred mass value is proportional to the time that the cell stayed occupied.
In this way, moving objects are differentiated from static or stopped objects.

\subsubsection{Fusion rule}
\label{fusion-rule}

An important part of the method consists in performing the fusion operation of a discounted and specialized \acl{PG} from preceding epoch ${}^{\alpha} m'_{PG, \, t-1}$
with a SG combined with prior knowledge from current epoch $m'{}_{SG,\, t}$.
The discounting operation is presented in section~\ref{sec:theory}
and the specialisation is described in the preceding paragraph.
In the section~\ref{prior}, combination of prior knowledge with the \acl{SG} is demonstrated.

\begin{equation}
    m_{PG,\, t} = {}^{\alpha}m'{}_{PG,\, t-1}\circledast m'{}_{SG,\, t}
    \label{eq:fusion-general}
\end{equation}

The fusion rule $\circledast$ is a modified conjunctive rule adapted to mobile object detection.
There are of course many different rules that could be used, but in order to distinguish between moving and stationary objects some modifications had to be performed.
These modifications consist in transferring the mass corresponding to a newly appeared object $\emptyset_{FO}$ to the class of moving objects $M$ as described by the equation \ref{eq:fusion-yager-modified}.
Symbol $\opconj$ denotes the conjunctive fusion rule.
\begin{align}
    \left(m_{1}\circledast m_{2}\right)\left(A\right) 
        &= \left(m_{1}\opconj m_{2}\right)(A)
        \nonumber \\
        & \forall A\subsetneq\Omega\wedge A\neq M
    \nonumber\\
    \left(m_{1}\circledast m_{2}\right)\left(M\right) 
        &= \left(m_{1}\opconj m_{2}\right)\left(M\right)+\left(m_{1}\opconj m_{2}\right)\left(\emptyset_{FO}\right)
    \nonumber \\
    \left(m_{1}\circledast m_{2}\right)\left(\Omega\right) 
        &= \left(m_{1}\opconj m_{2}\right)\left(\Omega\right)+\left(m_{1}\opconj m_{2}\right)\left(\emptyset_{OF}\right)
    \nonumber \\
    \left(m_{1}\circledast m_{2}\right)\left(\emptyset_{FO}\right)
        &= 0
    \nonumber \\
    \left(m_{1}\circledast m_{2}\right)\left(\emptyset_{OF}\right) 
        &= 0
    \label{eq:fusion-yager-modified}
\end{align}

All the above steps allow the construction of a \ac{PG} containing reach information on the environment state, including the knowledge on mobile and static objects.

\subsection{Fusion rule behaviour}
Proposed fusion scheme behaves differently depending on the context.
In this section, we describe briefly the behaviour of the fusion rule.
For an in-depth analysis, the reader is invited to read \cite{Kurdej2013ijar}.
\emph{Context} stands for prior knowledge information contained in \acl{GG}.
To demonstrate the effect of the fusion operator,
    we have chosen two particular cases, which clearly represent different contexts.

\subsubsection*{Building context}


In the building context, i.e. when $m(F) + m(I) + m(\Omega) \approx 1$, our fusion operator is roughly equivalent to the Yager's rule. 
The sum of conflict masses distinguished by the proposed rule is equal to the conflict mass in a regular fusion scheme without conflict management.
This behaviour is relevant, since it is assumed that no mobile obstacles are present in this context.
Therefore, only free space and infrastructure is to be distinguished.

\subsubsection*{Road and intermediate space}

%

%
The conflict management adapted to the perception scheme direct mass attribution to moving obstacles (class $M$).
The introduction of occupied space counter and \acl{PG} specialisation (see section~\ref{specialisation}) permits to transfer a part of the mass from \quotemarks{moving or other} class to \quotemarks{other}, where other is context-dependent.
\section{Experimental setup}
\label{sec:setup}

\subsection{Dataset}
\label{dataset}

The data set used for experiments was acquired in the 12th district of Paris.
The overall length of the trajectory was about 9~kilometres.
The vehicle pose comes from a system based on on a PolaRx II GPS and a NovAtel SPAN-CPT inertial measurement unit (IMU).
The system is supposed to provide precise positioning with high confidence.
Our main source of information about the environment is an IBEO Alaska XT lidar able to provide a cloud of about 800 points 10 times per second.
%
The digital maps that we use were provided by the \ac{IGN} and contain 3D building models as well as the road surface.
We also performed successful tests with freely available \emph{OpenStreetMap} project 2D maps \cite{OpenStreetMap},
but here we limited the use to building data.
We assume the maps to be accurate and up-to-date.


\subsection{\acl{GG} construction}
\label{gis-grid-construction}

The map data can be represented by two sets of polygons defining the 2D position of buildings and road surface by, respectively,
\begin{equation}
    \mathcal{B} =
    \left\lbrace
    b_{i} = 
        \begin{bmatrix}
            x_{1}x_{2}\ldots x_{m_{i}}
            \\
            y_{1}y_{2}\ldots y_{m_{i}}
        \end{bmatrix}
    , i\in[0,n_{B}]
    \right\rbrace
\end{equation}
%
%
\begin{equation}
    \mathcal{R} =
    \left\lbrace
    r_{i} = 
        \begin{bmatrix}
            x_{1}x_{2}\ldots x_{m_{i}}
            \\
            y_{1}y_{2}\ldots y_{m_{i}}
        \end{bmatrix}
    , i \in [0,n_{R}]
    \right\rbrace
\end{equation}

Our dataset satisfies the condition:
\(
    \mathcal{B} \cap \mathcal{R}=\emptyset
\).

We note that
    $B=\left\{ I\right\} $,
    $R=\left\{ F,\, S,\, M\right\} $,
    $T=\left\{ F,\, U,\, S,\, M\right\} $
    for convenience and readability only.
Set $A$ denotes then all other strict subsets of $\Omega$.
These aliases characterise the meta-information inferred from geographic maps.
For instance, on the road surface $R$, we $encourage$ the existence of free space $F$ as well as stopped $S$ and moving $M$ objects.
Analogically, building information $B$ fosters mass transfer to $I$.
Lastly, $T$ denotes the intermediate area, e.g. pavements, where mobile and stationary objects as well as small urban infrastructure can be present.
Please note that neither buildings nor roads are present,
    so the existence of mapped infrastructure $I$ can be excluded,
    but the presence of the other classes cannot.
Also, a level of confidence $\beta$ is defined for each map source, possibly different for each context.
Let $\tilde{x}=\frac{x_{-}+x_{+}}{2}$, $\tilde{y}=\frac{y_{-}+y_{+}}{2}$, then:

\begin{align}
    m_{GG}\{X,Y\}(B) & =\begin{cases}
        \beta_{B} & \text{if }(\tilde{x},\tilde{y})\in b_{i}\\
        0 & \text{otherwise}
    \end{cases}
    \\ \nonumber
    & \forall i\in[0,n_{B}]
    \\
    m_{GG}\{X,Y\}(R) & =\begin{cases}
        \beta_{R} & \text{if }(\tilde{x},\tilde{y})\in r_{i}\\
        0 & \text{otherwise}
    \end{cases}
    \\ \nonumber
    & \forall i\in[0,n_{R}]
\end{align}
\begin{align}
    m_{GG}\{X,Y\}(T) & =\begin{cases}
        0 & \text{if }(\tilde{x},\tilde{y})\in b_{i}\vee(\tilde{x},\tilde{y})\in r_{j}\\
        \beta_{T} & \text{otherwise}
    \end{cases}
    \\ \nonumber
    & \forall i\in[0,n_{B}],\forall j\in[0,n_{R}]
    \\
    m_{GG}\{X,Y\}(\Omega) & =\begin{cases}
        1-\beta_{B} & \text{if }(\tilde{x},\tilde{y})\in b_{i}\\
        1-\beta_{R} & \text{if }(\tilde{x},\tilde{y})\in r_{i}\\
        1-\beta_{T} & \text{otherwise}
    \end{cases}
    \\ \nonumber
    & \forall i\in[0,n_{B}],\forall j\in[0,n_{R}]
    \\
    m_{GG}\{X,Y\}(A) & =0
    \\ \nonumber
    & \forall A\subsetneq\Omega \text{ and } A\notin\left\{ B,R,T\right\} 
\end{align}

\subsection{Sensor model}

This section describes the way in which the data obtained from the sensor are transformed into the \acl{SG}.
If another exteroceptive sensor is used, one has to define an appropriate model.
The model used in the presented method is based on the one described in \cite{Moras2011}.

\subsection{Parameters}
\label{parameters}

The size of the grid cell in the occupancy grids was set to 0.5~m, which is sufficient to model a complex environment with mobile objects.
We have defined the map confidence factor $\mathbf{\beta}$ by ourselves, but ideally, it should be given by the map provider.
$\beta$ describes data currentness (age), errors introduced by geometry simplification and spatial discretisation.
$\mathbf{\beta}$ can also be used to depict the localisation accuracy.
Other parameters, such as counter steps $\delta_{inc}$, $\delta_{dec}$ and thresholds $\gamma_{O}$, $\gamma_{\emptyset}$ used for mobile object detection determine the sensitiveness of mobile object detection and were set by manual tuning.
Parameters used for the construction of \acl{SG},
    were set to $\mu_F = 0.7$, $\mu_O = 0.8$.

\section{Results}
\label{sec:results}


To assess the performance of our method,
    a comparison of perception results when prior knowledge from maps is present and when it is not available
    has been performed.
In this way, we show the interest of using a map-aided approach to the perception problem.

\newlength{\ImageHeight}
\setlength{\ImageHeight}{2.5cm}
\begin{figure*}[htb!]
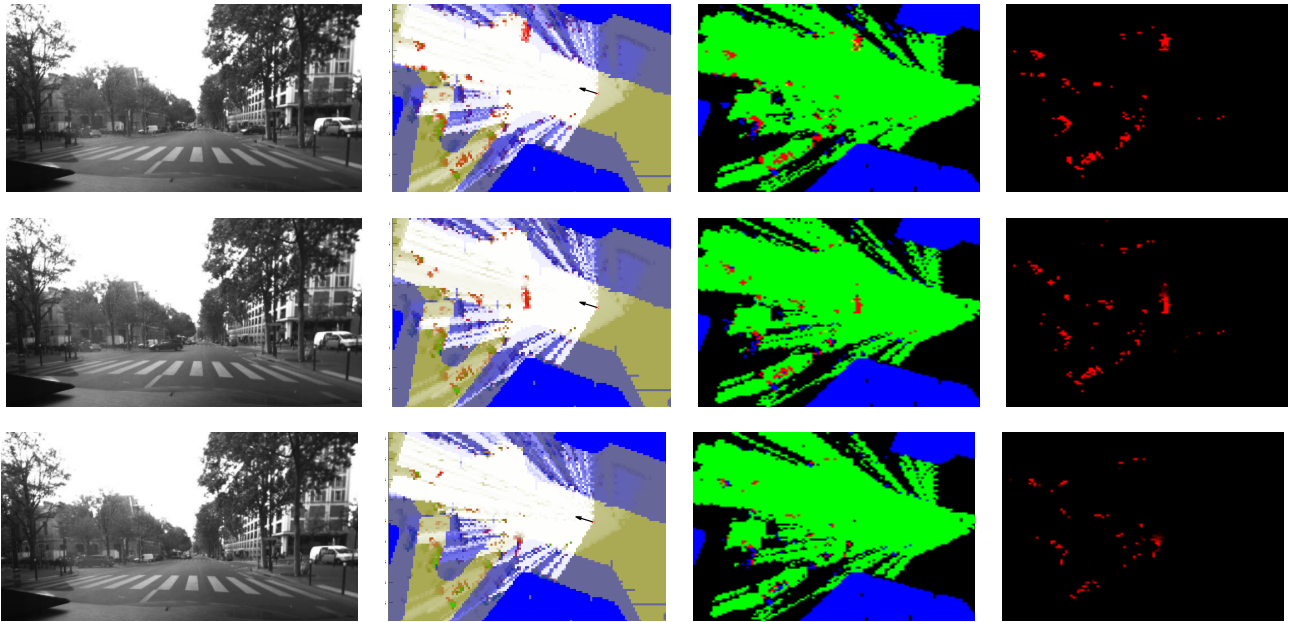

    \centering
    \foreach \imgno in {80,110,140}
    {
        \subfigure{
            \includegraphics[height=\ImageHeight,trim=0cm 5cm 0cm 0cm,clip]{figures/data/\padfivezeroes{\imgno}_image}
        }
        \subfigure{
            \includegraphics[height=\ImageHeight,width=0.21\textwidth,trim=2.5cm 2cm 7cm 2cm,clip]{figures/data/\padfivezeroes{\imgno}_betp}
        }
        \subfigure{
            \includegraphics[height=\ImageHeight,width=0.21\textwidth,trim=0 0 3cm 0,clip]{figures/data/\padfivezeroes{\imgno}_mixed}
        }
        \subfigure{
            \includegraphics[height=\ImageHeight,width=0.21\textwidth,trim=0 0 3cm 0,clip]{figures/data/\padfivezeroes{\imgno}_cumul}
        }
        \linebreak[2]
    }
    \caption{From left to right:
            (1) scene capture,
            (2) \acl{PG} pignistic probability,
            (3) simple decision rule to detect free space, moving and stopped obstacles,
            (4) trace of moving objects.
            Colour code for figures (3) and (4): green -- free space, red -- moving objects, blue -- static objects (buildings, stopped objects), black -- unknown space.
        }
    \label{fig:results}
\end{figure*}

The results for a particular instant of the approach tested on real-world data are presented on figure \ref{fig:results}.
The visualisation of the \ac{PG} has been obtained by attributing to each class a colour proportional to the pignistic probability $\betP$ and calculating the mean colour \cite{Smets2005}.
The presented scene contains two moving cars (only one is visible in the camera image) going in the direction perpendicular to the test
vehicle.

The principal advantage gained by using map knowledge is richer information on the detected objects.
A clear difference between a moving object (red, car) and a stopped objects (blue) is visible.
Also, stopped objects are distinct from infrastructure when prior map information is available (which is not highlighted on the figures).
In addition, thanks to the prior knowledge, stationary objects such as infrastructure are distinguished from stopped objects on the road.
Grids make noticeable the effect of discounting, as information on the environment behind the vehicle is being forgotten.

Figure~\ref{fig:results} shows also the effect of the discounting
    which is particularly visible on the free space behind the vehicle.
The grid cells get discounted, so the mass on the free class $F$ diminishes gradually.

\section{Conclusion and perspectives}
\label{sec:conclusion}


A new mobile perception scheme based on prior map knowledge has been introduced.
Geographic information is exploited to reduce the number of possible hypotheses delivered by an exteroceptive source.
A modified fusion rule taking into account the existence of mobile objects has been defined.
Furthermore, the variation in information lifetime has been modelled by the introduction of contextual discounting.


In the future, we anticipate removing the hypothesis that the map is accurate.
This approach will entail considerable work on creating appropriate error models for the data source.
Moreover, we envision differentiating the free space class into two complementary classes to distinguish navigable and non-navigable space.
This will be a step towards the use of our approach in autonomous navigation.
Another perspective is the use of reference data to validate the results, choose the most appropriate fusion rule and learn algorithm parameters.
We envision using map information to predict object movements.
It rests also a future work to exploit fully the 3D map information.


\section*{Acknowledgements}
This work was supported by the French Ministry of Defence DGA (Direction G\'{e}n\'{e}rale de l'Armement), with a Ph.D. grant delivered to Marek Kurdej.


\bibliographystyle{IEEEtran}
\bibliography{PhD-publications-conferences-iros2013}

\end{document}